\documentclass[journal]{IEEEtran}
\usepackage{times}
\usepackage{epsfig}
\usepackage{graphicx}
\usepackage{amsmath}
\usepackage{amssymb}
\usepackage{makecell}
\usepackage{booktabs}
\usepackage{array}
\usepackage{color}

\usepackage{cite}
\usepackage{xspace}
\usepackage[normalem]{ulem}
\useunder{\uline}{\ul}{}
\usepackage[table,xcdraw]{xcolor}
\usepackage{multirow}
\usepackage{threeparttable}
\usepackage{amsmath}
\usepackage[caption=false]{subfig}

\title{Anchor-free Oriented Proposal Generator for Object Detection}
\author{Gong Cheng, Jiabao Wang, Ke Li, Xingxing Xie, Chunbo Lang, Yanqing Yao, Junwei Han
\thanks{This work was supported in part by the National Science Foundation of China under Grant 62136007, Grant U20B2068, Grant 41871322, and Grant 42130112; in part by the Shaanxi Science Foundation for Distinguished Young Scholars under Grant 2021JC-16; and in part by the Fundamental Research Funds for the Central Universities. (\textit{Corresponding author: Ke Li.})}
\thanks{Gong Cheng, Jiabao Wang, Xingxing Xie, Chunbo Lang, Yanqing Yao, Junwei Han are with the School of Automation, Northwestern Polytechnical University, Xi'an 710129, China.}
\thanks{Ke Li is with the Zhengzhou Institute of Surveying and Mapping, Zhengzhou 450052, China (e-mail: like19771223@163.com).}
}

\begin{document}
\maketitle
\IEEEpeerreviewmaketitle

\begin{abstract}
Oriented object detection is a practical and challenging task in remote sensing image interpretation. Nowadays, oriented detectors mostly use horizontal boxes as intermedium to derive oriented boxes from them. However, the horizontal boxes are inclined to get small Intersection-over-Unions (IoUs) with ground truths, which may have some undesirable effects, such as introducing redundant noise, mismatching with ground truths, detracting from the robustness of detectors, etc. In this paper, we propose a novel Anchor-free Oriented Proposal Generator (AOPG) that abandons horizontal box-related operations from the network architecture. AOPG first produces coarse oriented boxes by a Coarse Location Module (CLM) in an anchor-free manner and then refines them into high-quality oriented proposals. After AOPG, we apply a Fast R-CNN head to produce the final detection results. Furthermore, the shortage of large-scale datasets is also a hindrance to the development of oriented object detection. To alleviate the data insufficiency, we release a new dataset on the basis of our DIOR dataset and name it DIOR-R. Massive experiments demonstrate the effectiveness of AOPG. Particularly, without bells and whistles, we achieve the accuracy of 64.41$\%$, 75.24$\%$ and 96.22$\%$ mAP on the DIOR-R, DOTA and HRSC2016 datasets respectively. Code and models are available at https://github.com/jbwang1997/AOPG.

\end{abstract}


\section{Introduction}

\IEEEPARstart{O}{riented} object detection, which exhibits a strong ability to analyze objects in remote sensing images, has attracted extensive attention from researchers. Apart from the function of generic detection, oriented object detection also requires the model to localize objects with oriented boxes so that the predictions can more accurately represent the positions of arbitrary-oriented instances. For the past few years, oriented object detection has made extensive progress~\cite{ding2019,xu2020,yang2019,han2021} by taking advantage of the deep learning algorithm. However, most of them are still along the route of Faster R-CNN~\cite{ren2016} to utilize horizontal anchors and proposals as references to generate oriented bounding boxes.

Referring to horizontal boxes as anchors or proposals to detect oriented objects suffers heavily from several drawbacks. (i) Horizontal boxes usually include massive background regions and multiple objects due to the shape inconsistency with arbitrary-oriented ground truths. This will introduce irrelevant information into region features, severely disturbing the network to obtain good performance. (ii) The Intersection-over-Unions (IoUs) between horizontal boxes and oriented ground truths are inclined to be small, which cannot explicitly reflect the matching degrees with ground truths. (iii) There is a huge gap, measured in terms of regression target, between horizontal proposals and oriented proposals. Since the regression targets of horizontal proposals are usually extremely large, it will hurt the robustness of the model.

\begin{figure}
	\begin{center}
	\includegraphics[width=\linewidth]{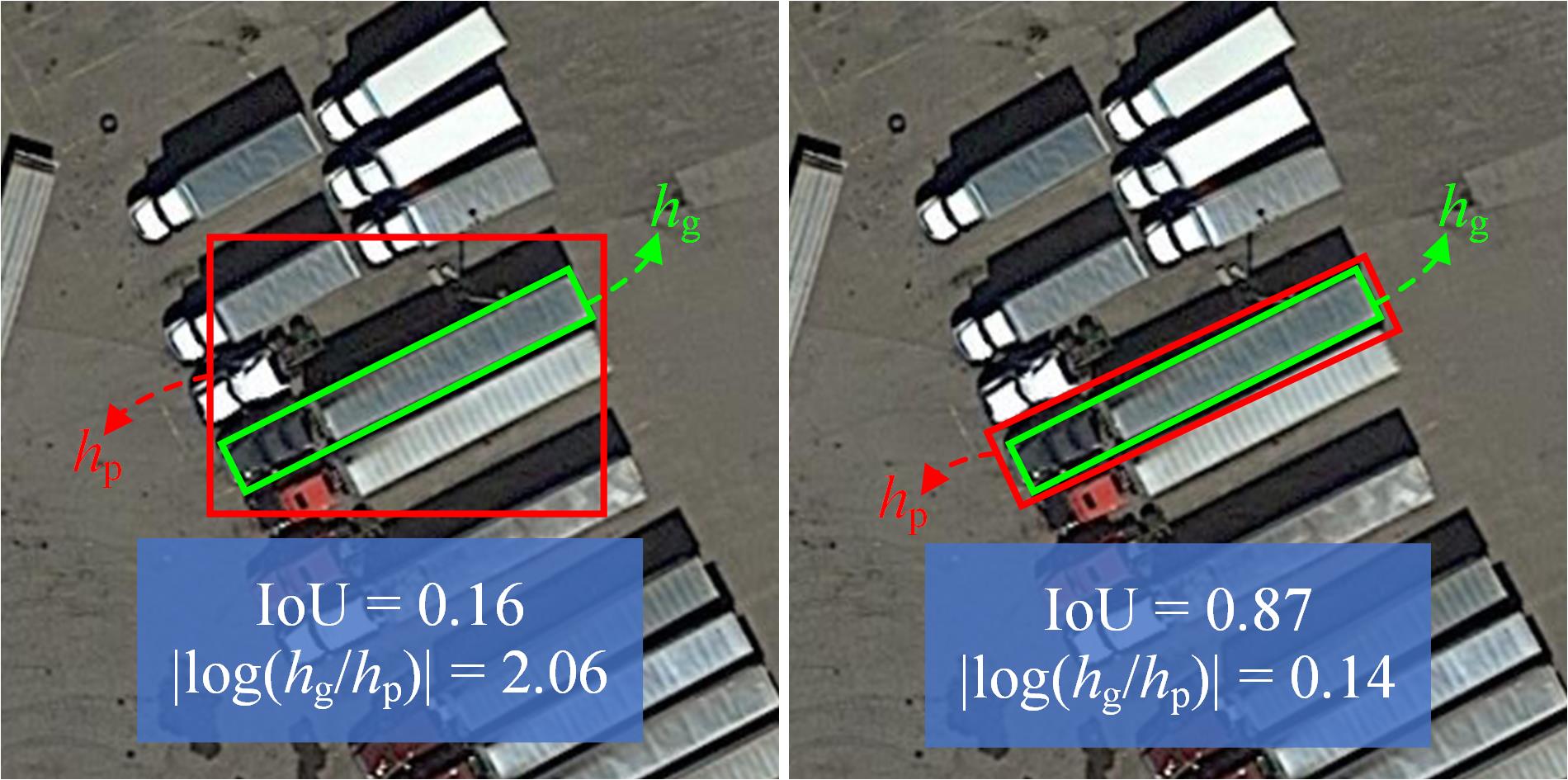}\\
	\end{center}
	\caption{Different cases of horizontal proposal and oriented proposal with the same ground truth. $h_{\mathrm{p}}$ denotes the height of the proposals and $h_{\mathrm{g}}$ denotes the height of the ground truths. As we can see, the horizontal proposal presents a small IoU and a big regression target measured in terms of $\left|\log \left(h_{\mathrm{g}} / h_{\mathrm{p}}\right)\right|$, while the oriented proposal has a large IoU and a low regression target.}
	\vspace*{-0.2cm}
	\label{img:issues}
\end{figure}

To illustrate the aforementioned problems, we show the horizontal and oriented proposals which are assigned with the same ground truth in Fig.~\ref{img:issues}. As we can see: (i) the horizontal proposal includes at least four vehicles, thus, it is very difficult for detectors to precisely classify and localize the targets. (ii) Even the horizontal proposal has been assigned as positive, it only has a small IoU of 0.16 with the ground truth. In contrast, the oriented proposal has a big IoU of 0.87. (iii) The regression target from the horizontal proposal to the ground truth reaches 2.06, which is very large for training detectors. As shown in the right figure, the regression target of the oriented proposal is just 0.14, which is much smaller than the horizontal proposal.

\begin{figure*}[h]
    \includegraphics{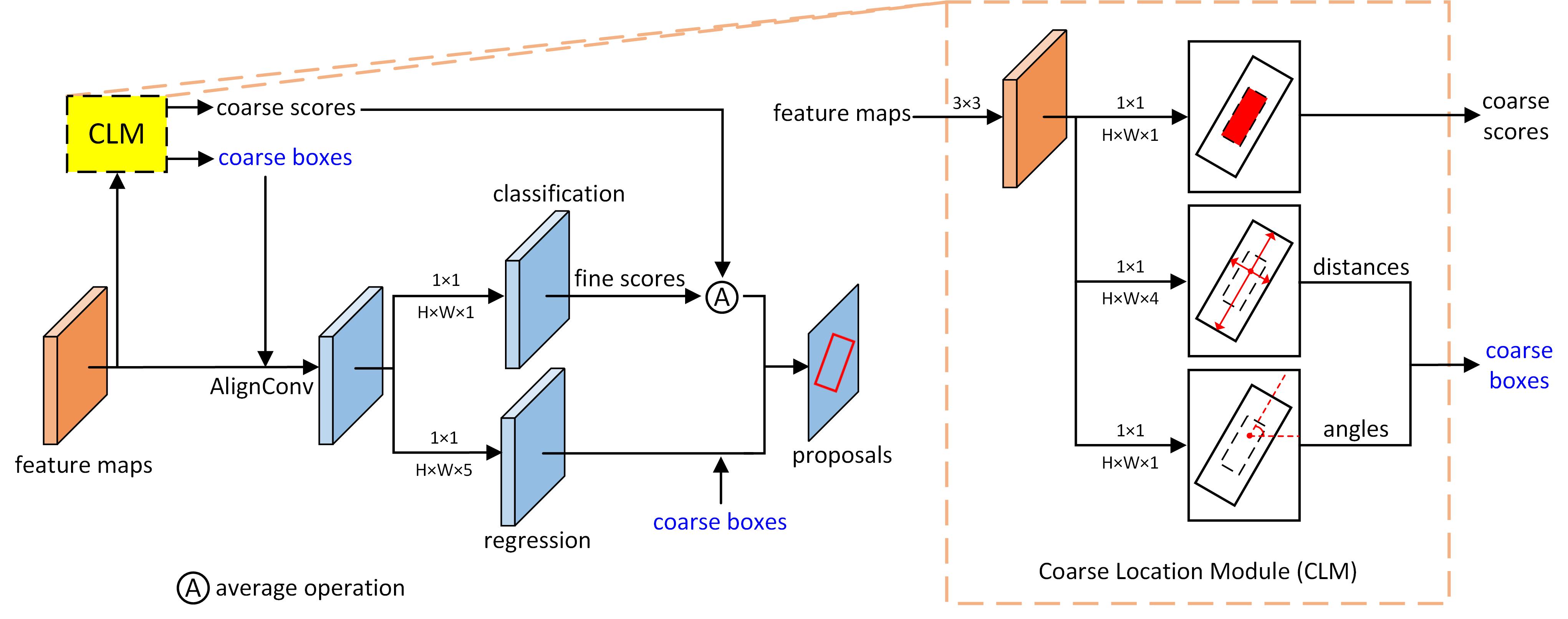}
    \caption{The architecture of AOPG. For each feature map in the feature pyramid,
           we use a Coarse Location Module (CLM) to predict the coarse scores, 
           distances, and angles of oriented boxes. Then we leverage the AlignConv 
           to eliminate the misalignment problem. Finally, a lightweight fully 
           convolutional network is applied to refine the coarse boxes to generate 
           high-quality proposals.}

   \label{img:structure}
\end{figure*}

In the past few years, many researchers have made tremendous effects to address those problems. For example, RoI Transformer~\cite{ding2019} attempts to train a lightweight module, named RoI Learner, to transform the original horizontal proposals into oriented ones. RoI Learner is flexible to plug in all two-stage detectors and significantly increases the detection accuracy. Motivated to alleviate feature misalignment, S$^2$ANet~\cite{han2021} first regresses oriented boxes based on horizontal anchors and then refines the features according to the predictions of the first stage. Next, more accurate oriented boxes are generated from the refined features. However, despite their success, those detectors exploit horizontal anchors as references to produce proposals. They still suffer heavily from the low IoUs and large shape distance problems brought by horizontal boxes. Although Rotated RPN~\cite{ma2018} straightly slides a bunch of oriented anchors on input images, it needs to place an excessive number of anchors in one pixel (a total of 54 oriented anchors with 3 scales, 3 ratios, and 6 angles) to cover all potential case of ground truths. The densely predefined anchors lead to big computation costs and detract from the model efficiency. What is more, manually designed oriented anchors still cannot match perfectly with ground truths.

To abolish all horizontal boxes in the model and generate high-quality oriented proposals, we deliver a novel Anchor-free Oriented Proposal Generator (AOPG) in this work. The main structure of our AOPG is shown in Fig.~\ref{img:structure}. Instead of sliding fixed-shape anchors on images, we adopt an anchor-free scheme to predict coarse oriented boxes at each position. Then, a novel technique named AlignConv is applied to align the features with coarse oriented boxes. After alignment, we further refine the coarse boxes into accurate locations to generate high-quality oriented proposals. In the second stage, we apply a Fast R-CNN head to predict the classification scores and regress the final oriented bounding boxes.

Furthermore, we observe that the shortage of large-scale datasets severely hinders the development of oriented object detection in remote sensing images. To alleviate the data insufficiency, we annotate the oriented bounding boxes for objects in our DIOR dataset~\cite{dior} and term it DIOR-R. Up to 23463 remote sensing images and 192512 instances covering 20 common classes are included in it.

We conduct comprehensive experiments to demonstrate the flexibility and effectiveness of our AOPG. Attributed to the high-quality oriented proposals, our AOPG achieves 64.41$\%$, 75.24$\%$ and 96.22$\%$ mAP without any bells and whistles on the DIOR-R, DOTA and HRSC2016 datasets.

In summary, our contributions lie in threefold as follows.
\begin{enumerate}
\item We analyze the drawbacks of using horizontal boxes in oriented object detection. To avoid those drawbacks, we propose an anchor-free oriented proposal generator. This novel design can produce high-quality proposals without involving horizontal boxes, which largely boosts detection accuracy.
\item We release a new large-scale oriented object detection dataset termed DIOR-R. There are up to 23463 remote sensing images and 192512 instances covering 20 common classes in this dataset. We also test several advanced methods on it and the results show that our DIOR-R is still challenging.
\item We conduct extensive experiments on the DIOR-R, DOTA and HRSC2016 datasets. The results demonstrate that our AOPG contributes to the improvement of detection accuracy. Particularly, our method obtains large gains over the baseline on all DIOR-R, DOTA and HRSC2016 datasets.
\end{enumerate}

\begin{figure*}[t]
    \includegraphics{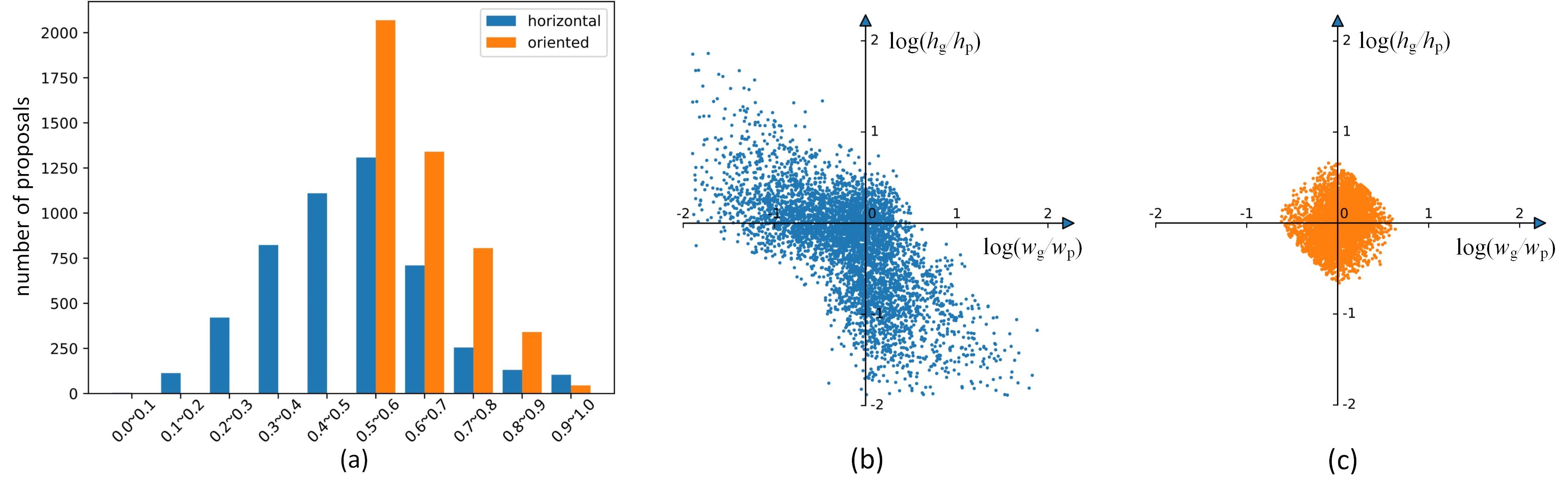}
	\centering
    \caption{Comparison of the horizontal-box scheme and the oriented-box scheme. 
            (a) is the IoU distributions of the horizontal proposals and the oriented proposals. (b) and (c) are the distributions of the regression targets 
            of the horizontal box-based detector and the oriented box-based detector, respectively.}

   \label{img:analyse}
\end{figure*}

\section{Related Work}

Recently, computer vision has rapidly grown with the help of deep learning technology~\cite{9328476,9314019,8252784,9435769,9239339,cheng2021perturbation}. As a crucial field of computer vision, object detection receives extensive attention from researchers and has made large progress in the past decades. Object detectors can be mainly separated into two paradigms based on their structures. Two-stage detectors, following the route of Faster R-CNN~\cite{girshick2015,ren2016,he2017mask}, first generate RoIs from a lightweight fully convolutional network and then extract the RoI features for precise classification and localization. Many latest works~\cite{liu2020deep,cai2018cascade,vu2019cascade,wang2019region,pang2019libra,lin2017feature,Cheng2021FeatureEN,9706434,li2017rotation,cheng2018learning} are based on the two-stage framework. Despite leading in many detection benchmarks, two-stage detectors face the problem of high computational complexity and complicated structure. On the contrary, one-stage detectors desire to directly complete detection by one fully convolutional network, which is more efficient and easier to implement. One-stage detectors often suffer from excessive negative samples since there is no sampling operation as in Faster R-CNN. To address this problem, Focal Loss~\cite{lin2017focal} dynamically gives a large weight for hard samples. With Focal Loss, one-stage detectors~\cite{redmon2016you,liu2016ssd,lin2017focal,zhang2020bridging} achieve competitive results with two-stage detectors. 

Benefiting from the development of generic detection, oriented object detection also has made great progress~\cite{xie2021oriented, ding2019,yang2019,xu2020,pan2020dynamic,yang2019r3det,cheng2016learning,xia2018dota,wang2019mask,fu2020point,wang2020learning,li2019feature} in recent years. SCRDet~\cite{yang2019} alleviates the influence of angle periodicity by designing a novel IoU-Smooth L1 Loss. R3Det~\cite{yang2019r3det} encodes centers and corners information in the features to get a more accurate location. Instead of predicting the angle of oriented boxes, Gliding Vertex~\cite{xu2020} predicts four-point polygons that are more explicit in images.

However, the majority of oriented detectors still use horizontal boxes as references. As discussed in the introduction, this scheme brings many drawbacks to oriented detectors. To avoid the problems brought by horizontal boxes, researchers have made lots of attempts. RoI Transformer~\cite{ding2019} trains RoI Learner module to transform horizontal proposals into oriented proposals, which significantly increases the detection accuracy. S$^2$ANet~\cite{han2021} uses the refinement stage to align oriented object features and produce more accurate predictions. Despite their success, those detectors exploit horizontal anchors as references to produce proposals. They still suffer heavily from the low IoUs and large shape distance problems brought by horizontal boxes. Some works~\cite{ma2018,liu2016ship} directly apply oriented anchors to images and then generate oriented RoIs or predict oriented objects. However, they need to put a massive number of oriented anchors on images to cover the huge oriented spatial domain. Thus, those detectors are more inefficient as well as face a serious positive and negative sample imbalance problem. What is more, as demonstrated in~\cite{tian2019fcos}, anchors need to be specifically designed in different datasets to achieve good results. Even with careful design, anchors still cannot predict some extreme samples.

In recent years, some works attempt to get rid of man-made anchors. Guided Anchoring~\cite{wang2019region} derives anchors from image features and then refines image features guided by the adaptive anchors. Cascade RPN~\cite{vu2019cascade}, following the Guided Anchoring route, runs a cascade structure to adapt anchors gradually.  FCOS~\cite{tian2019fcos} and FoveaBox~\cite{kong2020foveabox} directly regress the boxes from points. They predict distances from a point to the right, top, left, and bottom sides of objects, which totally abolish anchors. Different from the above works, CornerNet~\cite{law2018cornernet} and CenterNet~\cite{zhou2019objects} regard object detection as point detection~\cite{yang2017stacked}. CornerNet~\cite{law2018cornernet} detects and pairs the left-top and the right-bottom points of objects. CenterNet~\cite{zhou2019objects} directly predicts the center points of objects. In the aspect of oriented object detection, DRN~\cite{pan2020dynamic} also adopts the point detection strategy, which brings a new idea in oriented detection. BBAVector~\cite{yi2021oriented} adapts CenterNet~\cite{zhou2019objects} to the oriented object detector by taking advantage of box boundary-aware vectors.

In this work, we propose a novel Anchor-free Oriented Proposal Generator (AOPG) by adopting the anchor-free scheme towards generating high-quality oriented proposals. Instead of sliding horizontal anchors on the input images, we straightly generate oriented boxes from points. There is no horizontal box scheme in AOPG architecture, thus avoiding the problems brought by horizontal boxes. We will introduce the details of our model in the next section.

\section{Our Approach}
In this section, we first investigate the drawbacks of using horizontal box-based detection scheme for oriented object detection in section~\ref{sec:3.1} by comparing the proposals from the Faster R-CNN and our AOPG in two aspects. Next, we introduce the details of our AOPG. We adopt the Feature Pyramid Network (FPN)~\cite{lin2017feature} as the backbone. In each feature map, a Coarse Location Module (CLM) is applied to generate oriented boxes from feature points. The details of CLM will be presented in section~\ref{sec:3.2}. After receiving the coarse oriented boxes from CLM, we use the AlignConv operation to eliminate the misalignment between the features and the oriented boxes and then refine the coarse boxes using a small fully convolutional network. We simply average the coarse scores from the CLM and the fine scores from the fully convolutional network in order to get more accurate proposal scores. We will introduce the refinement process in section~\ref{sec:3.3}.

\subsection{Horizontal-Box Scheme V.S. Oriented-Box Scheme}
\label{sec:3.1}

Before introducing our model, we compare the horizontal-box scheme and oriented-box scheme for oriented object detection in two aspects, namely IoU distribution and regression target distribution. To this end, we train Faster R-CNN and our AOPG on the DOTA dataset, which are based on horizontal scheme and oriented scheme respectively. Firstly, we randomly collect 5000 positive proposals from each detector and calculate their IoUs with the ground truth boxes. The IoU distributions of the horizontal proposals and the oriented proposals are represented in Fig.~\ref{img:analyse} (a). As we can see, the majority of horizontal proposals have IoUs below 0.5 while all oriented proposals’ IoUs are higher than 0.5 thanks to the assignment with the oriented scheme. This proves that horizontal proposals cannot be precisely paired with ground truths. Additionally, horizontal proposals are inclined to have small IoUs with ground truths, which means that there exist background regions or irrelevant objects within them. This obviously decreases the accuracy of both classification and regression.

Besides, we calculate the regression targets of these two detectors and give their distributions measured in terms of the height and width target values for horizontal and oriented proposals, respectively, as shown in Fig.~\ref{img:analyse} (b) and Fig.~\ref{img:analyse} (c). It is clear that the targets of horizontal proposals are not symmetrically distributed around the origin. In addition,  there are many samples having extreme targets to regress. Those problems will increase the difficulty of training the detectors. In contrast, the targets of oriented proposals, as shown in Fig.~\ref{img:analyse} (c), are regular and small, which are much fitter for detectors to learn.

According to the comparison, we think that involving horizontal boxes in detectors will hurt the final accuracy of oriented object detection. So it is necessary to discard horizontal box-related operations in the detectors. This investigation motivates us to design AOPG, a novel framework to generate high-quality oriented proposals.

\subsection{Coarse Location Module}

\textbf{Oriented Box Definition.} As shown in Fig. 4, we define an oriented ground-truth box as $({{x}_{\text{gt}}},{{y}_{\text{gt}}},{{w}_{\text{gt}}},{{h}_{\text{gt}}},{{\theta }_{\text{gt}}})$. Here, $({{x}_{\text{gt}}},{{y}_{\text{gt}}})$ denotes its center coordinate, ${{w}_{\text{gt}}}$ and ${{h}_{\text{gt}}}$ represent its width and the height, and ${{\theta }_{\text{gt}}}$ stands for the clock-wise angle between its one side and the X-axis satisfying ${{\theta }_{\text{gt}}}\in \left[ {-\pi }/{4}\;,{\pi }/{4}\; \right]$. Given a positive sample indicated with $\left( x,y \right)$, its ground-truth distance vector respect to the left, top, right, bottom sides of ground-truth box is defined as ${{\boldsymbol{t}}_{\text{gt}}}=(l,t,r,b)$.

\label{sec:3.2}

\begin{figure}[t]
    \includegraphics[width=2.7in]{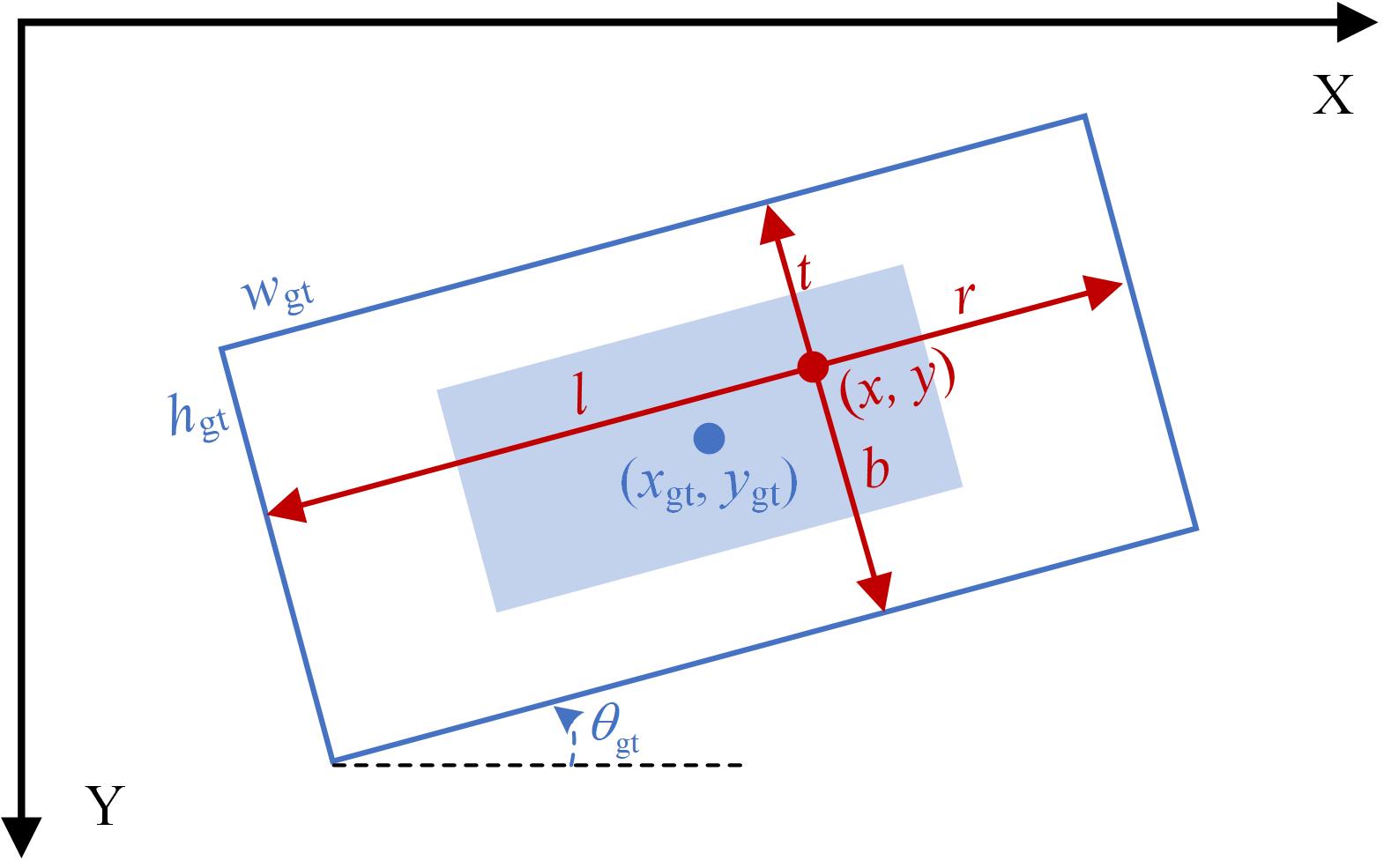}
	\centering
    \caption{Oriented box definition in our AOPG. The blue box is ground truth. 
            The points located in the blue region are positives.}
    \label{img:box_defination}
\end{figure}

\textbf{Region Assignment.} Anchor-based detectors use IoUs between anchors and ground truths to assign samples as positives or negatives. However, our CLM regresses oriented boxes from points straightly, so there is no anchor in the model. Thus, we apply a region assignment scheme to replace the traditional IoU assignment. In the training phase, each ground-truth box will be first assigned to one feature map of FPN according to its box size. Next, we project the feature maps back onto the images and choose the points that locate in the central regions of ground-truth boxes as positive samples and the others are negative samples.

Specifically, we make use of five levels of feature maps defined as $\{{{P}_{2}},{{P}_{3}},{{P}_{4}},{{P}_{5}},{{P}_{6}}\}$, where their strides $\{{{s}_{2}},{{s}_{3}},{{s}_{4}},{{s}_{5}},{{s}_{6}}\}$ are 4, 8, 16, 32, and 64, respectively. The ground truth with the size belonging in $[{{{\alpha }^{2}}s_{i}^{2}}/{2}\;,2{{\alpha }^{2}}s_{i}^{2}]$ is assigned to ${{P}_{i}}$, where $\alpha $ is a factor to scale up the range. Especially, we set the minimum ground-truth size assigned to ${{P}_{2}}$ as 0 and the maximum ground-truth size assigned to ${{P}_{6}}$ as 100000 in order to cover all different sizes of ground truths.

After assigning each ground truth to its corresponding feature map, we label the points as positives if they lie in the central regions of ground truths. The central region of a ground truth can be expressed as $\boldsymbol{B}_{\sigma }^{{}}=(x_{\text{gt}}^{{}},y_{\text{gt}}^{{}},\sigma w_{\text{gt}}^{{}},\sigma h_{\text{gt}}^{{}},\theta _{\text{gt}}^{{}})$, where $\sigma $ is the central rate. In order to judge whether a point $(x_{\text{pt}}^{{}},y_{\text{pt}}^{{}})$ on ${{P}_{i}}$ locates in the $\boldsymbol{B}_{\sigma }^{{}}$, we need to convert the point from the image coordinate system to its ground-truth coordinate system by

\begin{equation}
\left( \begin{matrix}
   {{x}'}  \\
   {{y}'}  \\
\end{matrix} \right)=\left( \begin{matrix}
   \cos {{\theta }_{\text{gt}}} & -\sin {{\theta }_{\text{gt}}}  \\
   \sin {{\theta }_{\text{gt}}} & \cos {{\theta }_{\text{gt}}}  \\
\end{matrix} \right)\left( \begin{matrix}
   x-x_{\text{gt}}^{{}}  \\
   y-y_{\text{gt}}^{{}}  \\
\end{matrix} \right).
\end{equation}

If the coordinate of a point after transformation satisfies $\left| {{x}'} \right|<{\sigma w_{\text{gt}}^{{}}}/{2}\;\text{ and }\left| {{y}'} \right|<{\sigma h_{\text{gt}}^{{}}}/{2}\;$, the point lies in the central region of the ground truth. Thus, it is a positive sample, as shown in Fig.~\ref{img:box_defination}.

\textbf{Training.} As shown in Fig.~\ref{img:structure}, our CLM has three branches, which produce coarse boxes’ scores, distances, and angles in each position, represented by $c$, $\boldsymbol{t}$, and $\theta $. 

The score branch is a classification branch to indicate the central region of ground truths. We train this branch with all points obtained with the aforementioned region assignments.

The distance branch predicts the distances from each point on the feature map to the left, top, right, bottom sides of ground-truth boxes. We only train the distance branch on positive samples. Because the ground-truth coordinate system is unparallel with the image coordinate system, we need to transform each point on the feature map to its corresponding ground-truth coordinate system in the same way as region assignment. Next, as we can see in Fig.~\ref{img:box_defination}, the ground-truth distance vector ${{\boldsymbol{t}}_{\text{gt}}}=(l,t,r,b)$ of a point with coordinate index $(x,y)$ can be formulated as 
\begin{equation}    
   \left\{ \begin{aligned}
  & l={{{w}_{\text{gt}}}}/{2}\;+{x}'\text{,  }r={{{w}_{\text{gt}}}}/{2}\;-{x}' \\ 
 & t={{{h}_{\text{gt}}}}/{2}\;+{y}'\text{,  }b={{{h}_{\text{gt}}}}/{2}\;-{y}' \\
\end{aligned} \right.
.
\end{equation}
We train the distance branch by the normalized distance vector $\boldsymbol{t}_{\text{gt}}^{*}=({{l}^{*}},{{t}^{*}},{{r}^{*}},{{b}^{*}})$, where $\boldsymbol{t}_{\text{gt}}^{*}=\log ({{{\boldsymbol{t}}_{\text{gt}}}}/{z)}\;$, and $z$ is a normalizing factor defined in each feature map.

\begin{figure}[t]
    \includegraphics[width=3.0in]{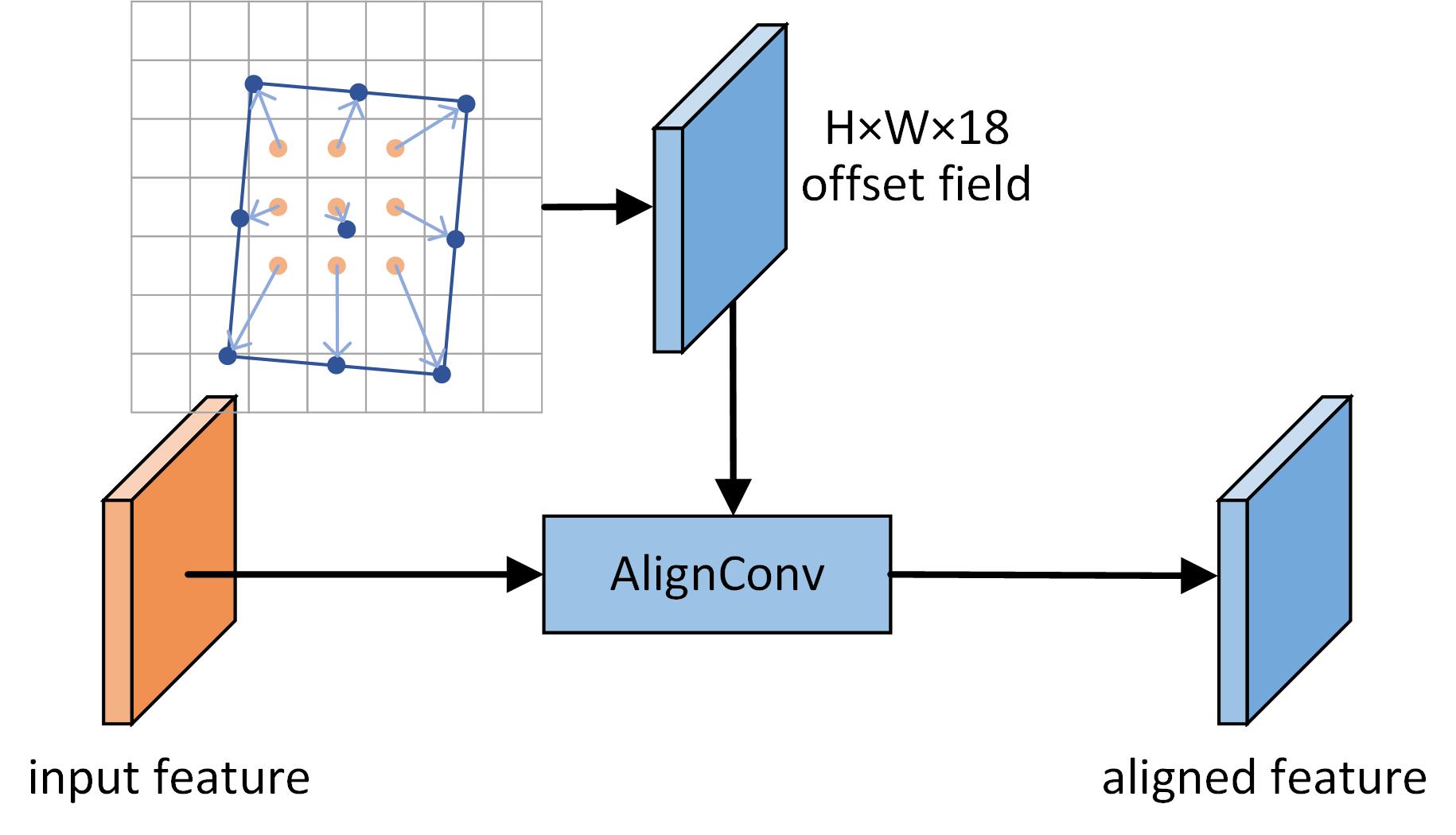}
    \centering
    \caption{The architecture of AlignConv. The offset field is calculated according to the oriented 
            box at each position.}
    \vspace*{-0.2cm}
    \label{img:alignconv}
\end{figure}

In the definition of oriented boxes, the angle is in a symmetric interval, where ${{\theta }_{\text{gt}}}\in \left[ {-\pi }/{4}\;,{\pi }/{4}\; \right]$. Therefore, we directly use the ground-truth angles as the targets to train the angle branch. Thus, the loss function of CLM can be formulated as
\begin{equation}
\begin{small}
   \begin{aligned}
  & {{L}_{\text{CLM}}}=\frac{\lambda }{N}\sum\limits_{(x,y)}{{{L}_{\text{ctr}}}(c,c_{\text{gt}}^{*})}+\frac{1}{{{N}_{\text{pos}}}}\sum\limits_{(x,y)}{c_{\text{gt}}^{*}{{L}_{\text{dist}}}(\boldsymbol{t},\boldsymbol{t}_{\text{gt}}^{*})} \\ 
 & +\frac{1}{{{N}_{\text{pos}}}}\sum\limits_{(x,y)}{c_{\text{gt}}^{*}{{L}_{\text{angle}}}}(\theta ,{{\theta }_{\text{gt}}}), \\ 
\end{aligned}
\end{small}
\end{equation}
where ${{L}_{\text{ctr}}}$ is the focal loss as in~\cite{lin2017focal}, ${{L}_{\text{dist}}}$ and ${{L}_{\text{angle}}}$ are the smooth L1 loss as in~\cite{ren2016}. $N$ denotes the number of all points on feature maps and ${{N}_{\text{pos}}}$ is the number of positive points. $c_{\text{gt}}^{*}$ indicates the point labels obtained through the region assignment scheme that $c_{\text{gt}}^{*}$ is 1 when a point is positive, otherwise 0.

By combining distances and angles, CLM can generate one oriented box in each location. The comparison of the coarse oriented boxes with the horizontal proposals generated by RPN is made in Section IV.C, which shows that the coarse oriented boxes are much better than human-made anchors.

\subsection{Oriented Box Refinement}
\label{sec:3.3}

Next, we apply a tiny fully convolutional network to identify the foreground and precisely refine the coarse oriented boxes. However, the anchors are uniform on the whole feature map, and they share the same shape and scale in each position. Our coarse oriented boxes vary across locations, which have a misalignment matter with consistent feature maps. In this work, we apply a feature alignment technique called AlignConv to align the features with the coarse oriented boxes. After alignment, high-quality oriented proposals are generated for precise classification and location.

\textbf{Feature Alignment by AlignConv.} Fig.~\ref{img:alignconv} shows the framework of AlignConv. The main part of AlignConv is a deformable convolution layer, which aligns the input feature guided by the offset field. Different from normal deformable convolution that generates the offset field from a small network, AlignConv derives the offset field from oriented boxes. To be more specific, for a position vector $\boldsymbol{p}\in \left\{ 0,1,...,H-1 \right\}\times \left\{ 0,1,...,W-1 \right\}$, a standard 3$\times$3 deformable convolution operation can be represented as  
\begin{equation}
   \mathbf{Y}(\boldsymbol{p})=\sum\limits_{\boldsymbol{r}\in \mathcal{R}}{\mathbf{W}(\boldsymbol{r})\cdot \mathbf{X}(\boldsymbol{p}+\boldsymbol{r}+\boldsymbol{o})},
\end{equation}
where $\mathbf{X}$ and $\mathbf{Y}$ are the input and output features, $\mathbf{W}$ is the kernel weights of deformable convolution, $\boldsymbol{r}$ is a vector element from the regular grid $\mathcal{R}=\left\{ (-1,-1),(-1,0),...,(1,1) \right\}$ and $\boldsymbol{o}$ denotes the position offsets. In AlignConv, we constrain the sampling points to obey a regular distribution in the coarse oriented box $\boldsymbol{B} =(x,y,w,h,\theta )$ as shown in Fig.~\ref{img:alignconv}. The sampling position can be deduced by
\begin{equation}
   {{\boldsymbol{r}}_{\text{box}}}=\frac{1}{{{s}_{i}}}\left( \left( x,y \right)-\boldsymbol{p}+(\frac{w}{2},\frac{h}{2})\cdot 
   \boldsymbol{r} \right){{\mathbf{R}}^{\mathbf{T}}}\left( \theta  \right),
\end{equation}
where $\mathbf{R}(\theta )$ is the rotation matrix, the same as~\cite{han2021} and ${{s}_{i}}$ is the stride of feature map ${{P}_{i}}$. We can calculate the offset field $\mathcal{O}$ in position $\boldsymbol{p}$ by
\begin{equation}
\mathcal{O}=\sum\limits_{\boldsymbol{r}\in \mathcal{R}}{\left( {{\boldsymbol{r}}_{\text{box}}}-\boldsymbol{p}-\boldsymbol{r} \right)}.
\end{equation}

\begin{figure}[t]
    \includegraphics[width=3.0in]{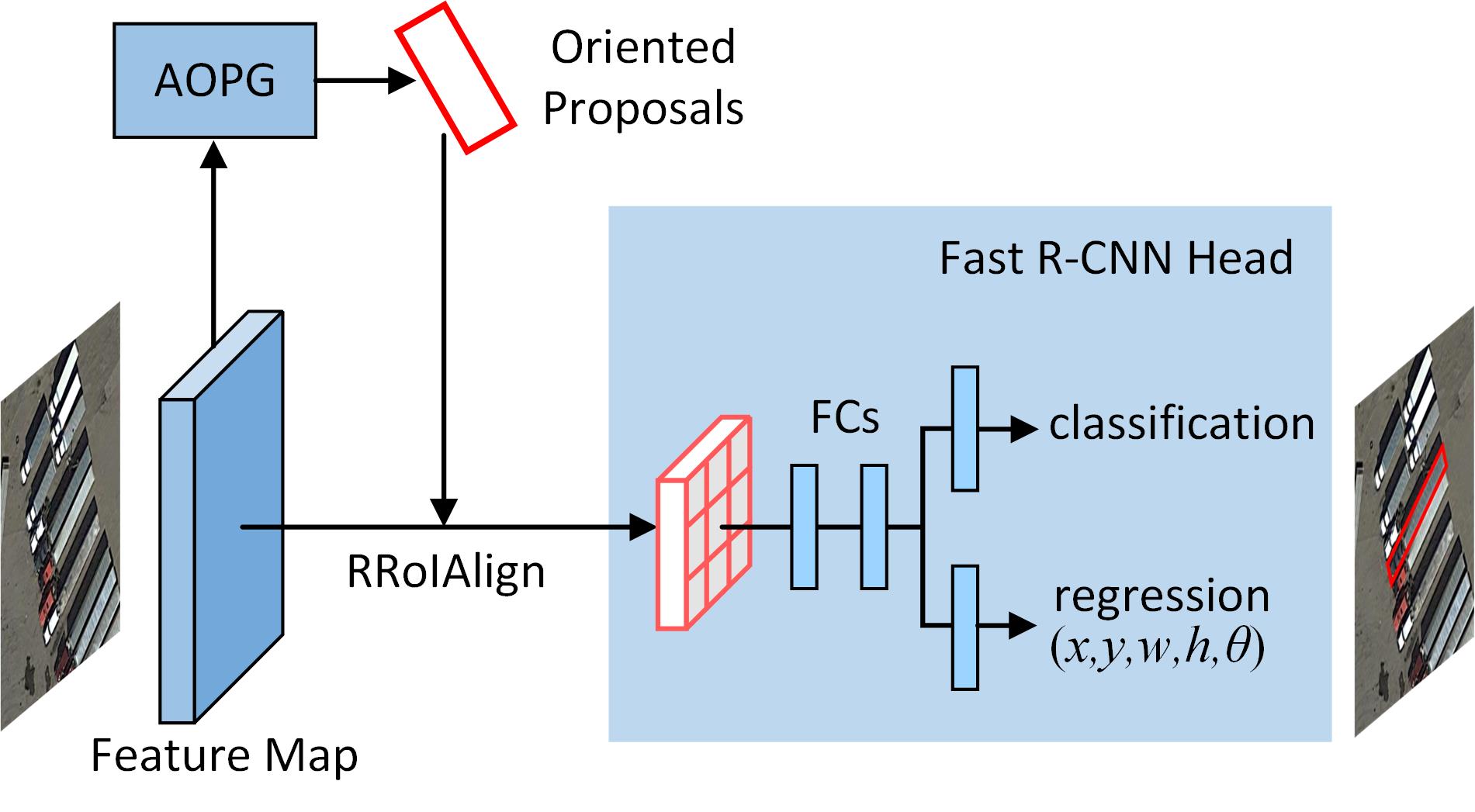}
    \centering
    \caption{The structure of Fast R-CNN head for object detection.}
    \vspace*{-0.1cm}
    \label{img:fast_rcnn}
\end{figure}

\begin{figure*}[t]
	\begin{center}
		\includegraphics[width=0.95\linewidth]{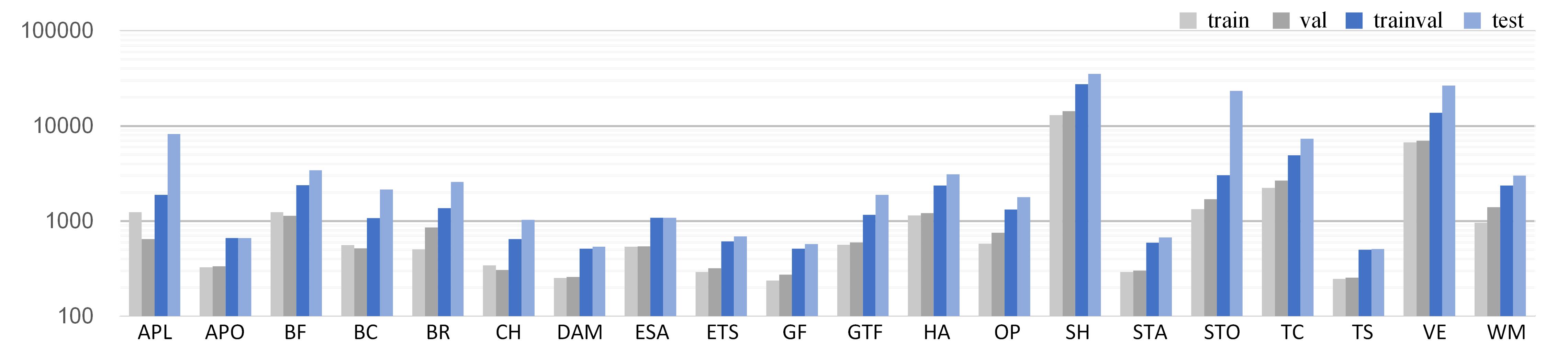}\\
	\end{center}
	\vspace*{-0.4cm}
	\caption{The instance numbers of different categories and sets.}\label{img:static}
\end{figure*}

\begin{figure*}[t]
	\begin{center}
		\includegraphics[width=0.95\linewidth]{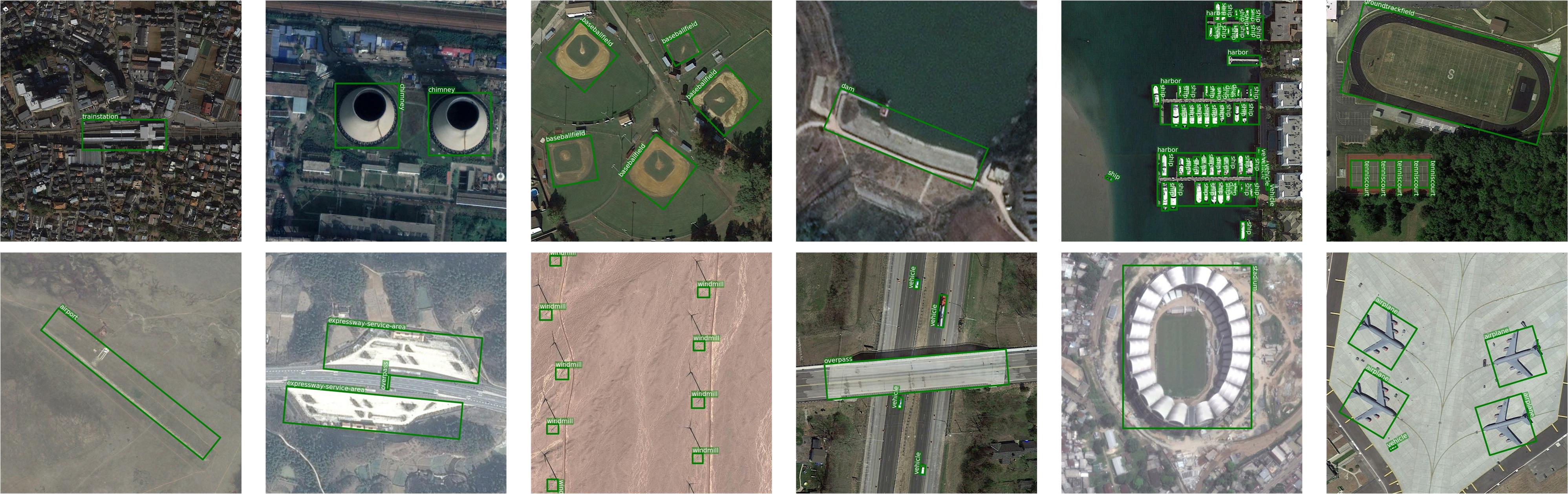}\\
	\end{center}
	\vspace*{-0.4cm}
	\caption{Some examples in the DIOR-R dataset.}\label{img:dior_label}

\end{figure*}

\textbf{High-quality Oriented Proposals.} After feature alignment, we apply two 1$\times$1 convolution layers to generate classification and regression maps, separately, targeting to identify the foreground and refine the oriented boxes. For proposal refinement, we need to first calculate the coarse boxes’ IoUs with ground-truth boxes to assign their labels. Here, the coarse boxes are generated via CLM. The boxes with the IoUs higher than 0.7 are considered as positives and the ones lower than 0.3 are regarded as negatives. The others are ignored. With the definitions of positive and negative samples, the coarse boxes are refined into accurate locations to generate high-quality oriented proposals. 

\subsection{Oriented Object Detection} 
As shown in Fig.~\ref{img:fast_rcnn}, we adopt modified Fast R-CNN head in the second stage to predict the classification scores and regress the final oriented bounding boxes. Different from original Fast R-CNN head, we add an angle parameter on the regression branch for angle bias prediction. The details of AOPG can be referred in our released code. \footnote{https://github.com/jbwang1997/AOPG}

\section{Experiments}
\subsection{Datasets and Training Details}
\textbf{Datasets.} DOTA\footnote{https://captain-whu.github.io/DOTA/dataset.html}~\cite{xia2018dota} is a large-scale aerial object detection dataset containing 2806 images and 188282 instances of 15 common object classes: Harbor (HA), Storage Tank (ST), Large Vehicle (LV), Swimming Pool (SP), Soccer Ball Field (SBF), Bridge (BR), Ship (SH), Plane (PL), Baseball Diamond (BD), Tennis Court (TC), Ground Track Field (GTF), Basketball Court (BC), Small Vehicle (SV), Roundabout (RA) and Helicopter (HC). We use both training and validation sets for training and evaluate the model on the testing set. All results are obtained through the DOTA evaluation server.

HRSC2016\footnote{https://sites.google.com/site/hrsc2016/annotations}~\cite{liu2016ship} is a challenging ship detection dataset including 1061 images with the size ranging from 300$\times$300 to 1500$\times$900. In the experiments, both the training set (436 images) and validation set (181 images) are used for training. We evaluate the models at the testing set (444 images) in terms of PASCAL VOC07 and VOC12 metrics.

\begin{figure*}
	\begin{center}
		\vspace{-0.1cm}
		\includegraphics[width=0.95\linewidth]{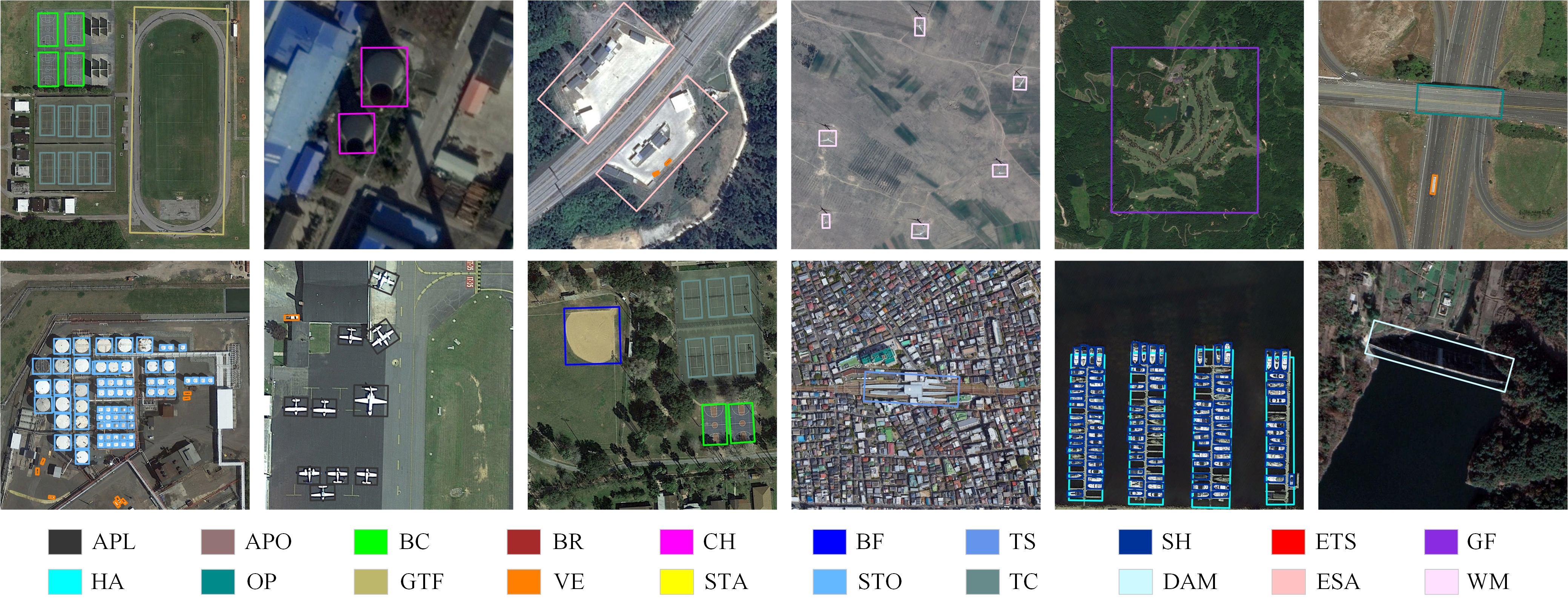}\\
	\end{center}
	\vspace{-0.4cm}
	\caption{The visualization of the detection results on the DIOR-R dataset.}\label{img:dior_vis}
\end{figure*}

\begin{figure*}
	\begin{center}
		\vspace{-0.1cm}
		\includegraphics[width=0.95\linewidth]{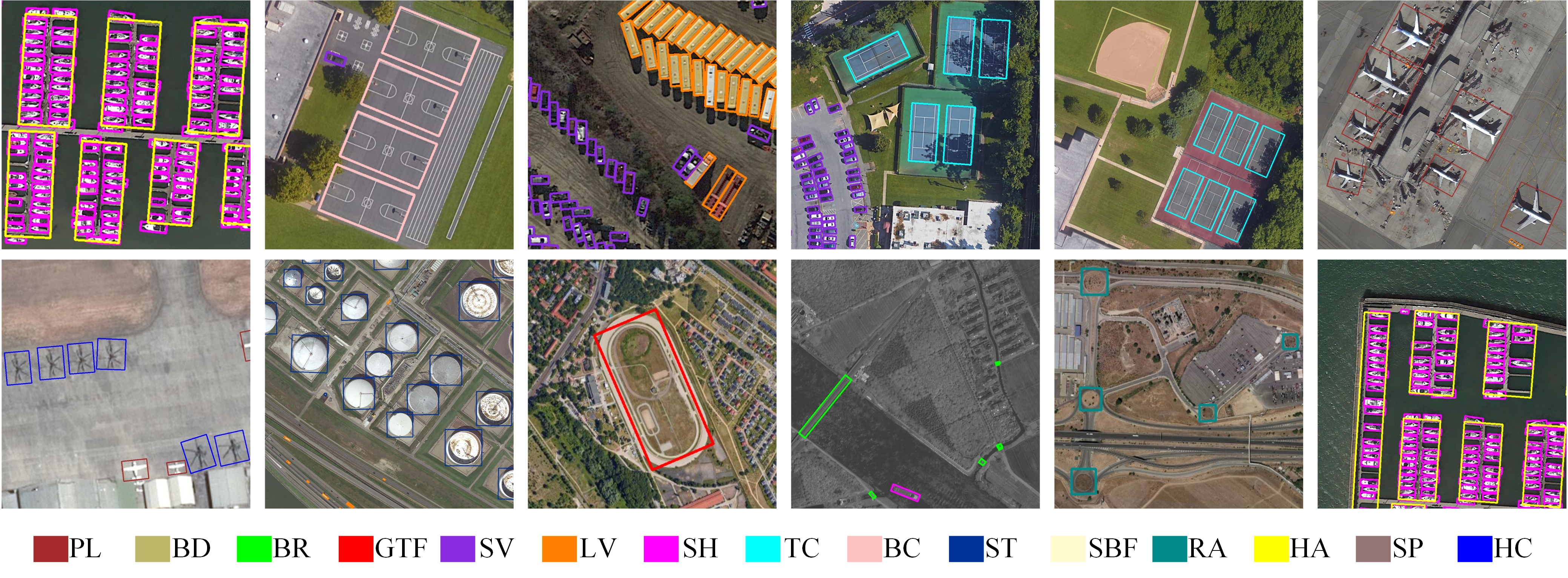}\\
	\end{center}
	\vspace{-0.4cm}
	\caption{The visualization of the detection results on the DOTA dataset.}\label{img:dota_vis}
\end{figure*}

\begin{figure*}
	\begin{center}
		\vspace{-0.1cm}
		\includegraphics[width=0.95\linewidth]{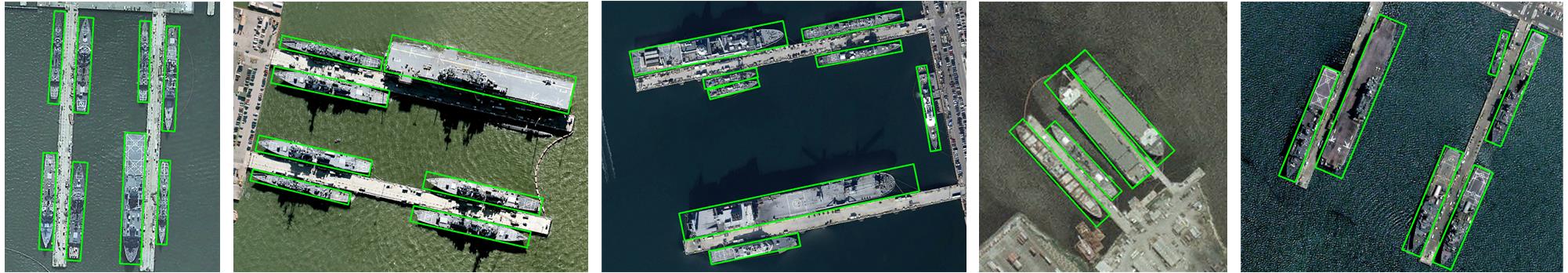}\\
	\end{center}
	\vspace*{-0.4cm}
	\caption{The visualization of the detection results on the HRSC2016 dataset.}\label{img:hrsc_vis}
\end{figure*}

In addition to the above datasets, we also release a new oriented object detection dataset termed DIOR-R\footnote{https://gcheng-nwpu.github.io}. Our DIOR-R shares the same images with the previous DIOR dataset~\cite{dior}. Differently, we annotate the oriented bounding boxes for every instance to adapt to the oriented detection task. There are 23463 images and 192518 instances in this dataset, covering a wide range of scenes and 20 common object classes. The categories of objects in DIOR-R include Airplane (APL), Airport (APO), Baseball Field (BF), Basketball Court (BC), Bridge (BR), Chimney (CH), Expressway Service Area (ESA), Expressway Toll Station (ETS), Dam (DAM), Golf Field (GF), Ground Track Field (GTF), Harbor (HA), Overpass (OP), Ship (SH), Stadium (STA), Storage Tank (STO), Tennis Court (TC), Train Station (TS), Vehicle (VE) and Windmill (WM). We count the instance numbers of different categories in Fig.~\ref{img:static} and visualize some examples in Fig.~\ref{img:dior_label}.

\textbf{Training Details.} Unless specified, we adopt ResNet50~\cite{he2016identity} with FPN~\cite{lin2017feature} as the backbone and use the same hyper-parameters with oriented Faster R-CNN. Note that we set the factors $\alpha$ and $\sigma$  in region assignment to 8 and 0.2 respectively. The normalizing factors $z$ are 16, 32, 64, 128, 256, which correspond to the feature levels $P_{2}, P_{3}, P_{4}, P_{5}$ and  $P_{6}$. The NMS threshold for proposal generation is set to 0.8. All experiments are conducted on one NVIDIA RTX 2080 Ti with a batch size of 2. In the training phase, we load the model pretrained on the ImageNet~\cite{deng2009imagenet}. We optimize the model by Stochastic Gradient Descent (SGD) algorithm with the initial learning rate of 0.005, the momentum of 0.9 and the weight decay of 0.0001. Random horizontal and vertical flipping have been applied during training to avoid over-fitting.

The image size of the DOTA dataset ranges from 800$\times$800 to 4000$\times$4000, which is too big for training. Following~\cite{ding2019}, we crop original images into a series of 1024$\times$1024 patches with the stride of 824. For multi-scale experiments, we first resize original images into three scales (0.5, 1.0, 2.0) and then crop all the images of these three scales into 1024$\times$1024 with a stride of 524. We run 12 epochs and reduce the learning rate by a factor of 10 at the end of epoch 8 and epoch 11. For the DIOR-R dataset, we keep the image size at the original size of 800$\times$800. The same training schedule as the DOTA dataset is implemented. For the HRSC2016 dataset, we resize the images to the (800, 1333) without changing the aspect ratios. We run 36 epochs and reduce the learning rate by a factor of 10 at epoch 24 and epoch 33. Our model is based on the mmdetection~\cite{chen2019mmdetection} library, which is an open-source object detection toolbox based on PyTorch.

\begin{center} 
\begin{table*}
\renewcommand\arraystretch{1.5}
\caption{Comparison with advanced methods on the DIOR-R dataset.}
\vspace{-0.1cm}
\tiny
\setlength\tabcolsep{4.5pt}
\begin{tabular}{c|c|c|c|c|c|c|c|c|c|c|c|c|c|c|c|c|c|c|c|c|c|c}
\hline
Methods        & Backbone & APL             & APO             & BF             & BC             & BR             & CH             & DAM             & ETS             & ESA             & GF            & GTF            & HA         & OP            & SH            & STA            & STO            & TC            & TS            & VE            & WM           & mAP           \\ \hline
Faster RCNN-O~\cite{ren2016}  & R-50-FPN & 62.79          & 26.80           & 71.72          & 80.91          & 34.20           & 72.57          & 18.95          & 66.45          & 65.75          & 66.63          & 79.24          & 34.95       & 48.79          & 81.14          & 64.34          & 71.21          & 81.44          & 47.31          & 50.46          & 65.21         & 59.54         \\
RetinaNet-O~\cite{lin2017focal}    & R-50-FPN & 61.49          & 28.52          & 73.57          & 81.17          & 23.98          & 72.54          & 19.94          & 72.39          & 58.20           & 69.25          & 79.54          & 32.14       & 44.87          & 77.71          & 67.57          & 61.09          & 81.46          & 47.33          & 38.01          & 60.24         & 57.55         \\
Gliding Vertex~\cite{xu2020} & R-50-FPN & \textbf{65.35} & 28.87          & \textbf{74.96} & 81.33          & 33.88          & \textbf{74.31} & 19.58          & 70.72          & 64.70           & 72.30           & 78.68          & 37.22       & 49.64          & 80.22          & 69.26          & 61.13          & 81.49          & 44.76          & 47.71          & 65.04         & 60.06         \\
RoI Transformer~\cite{ding2019}      & R-50-FPN & 63.34          & \textbf{37.88}          & 71.78          & 87.53 & 40.68          & 72.60           & 26.86          & \textbf{78.71}          & 68.09          & 68.96          & \textbf{82.74} & \textbf{47.71}       & \textbf{55.61}          & \textbf{81.21}          & \textbf{78.23} & 70.26          & \textbf{81.61} & 54.86          & 43.27          & 65.52         & 63.87         \\ \hline
\textbf{AOPG}  & R-50-FPN & 62.39           & 37.79 & 71.62          & \textbf{87.63}          & \textbf{40.90} & 72.47          & \textbf{31.08} & 65.42 & \textbf{77.99} & \textbf{73.20} & 81.94          & 42.32 & 54.45 & 81.17 & 72.69          & \textbf{71.31} & 81.49          & \textbf{60.04} & \textbf{52.38} & \textbf{69.99} & \textbf{64.41} \\ \hline
\end{tabular}
\label{table:dior_comparison}
\end{table*}
\end{center}

\begin{table*}\small
	\setlength{\belowcaptionskip}{-1cm}
	\renewcommand\arraystretch{1.5}
	\caption{Comparison with state-of-the-art methods on the DOTA dataset. * means the results from AerialDetection (the same below). $^\ddagger$ denotes multi-scale training and testing.}
	\vspace{-0.6cm}
	\label{table:dota_comparison}
	\begin{center}
		\resizebox{1.0\textwidth}{!}{
			\begin{tabular}{p{3cm}<{\centering}|p{2.5cm}<{\centering}|p{0.8cm}<{\centering}|p{0.8cm}<{\centering}|p{0.8cm}<{\centering}|p{0.8cm}<{\centering}|p{0.8cm}<{\centering}|p{0.8cm}<{\centering}|p{0.8cm}<{\centering}|p{0.8cm}<{\centering}|p{0.8cm}<{\centering}|p{0.8cm}<{\centering}|p{0.8cm}<{\centering}|p{0.8cm}<{\centering}|p{0.8cm}<{\centering}|p{0.8cm}<{\centering}|p{0.8cm}<{\centering}|p{0.8cm}<{\centering}|p{0.8cm}<{\centering}}
				\hline
				Method                      & Backbone  & PL             & BD             & BR             & GTF            & SV             & LV             & SH             & TC             & BC             & ST             & SBF            & RA             & HA             & SP             & HC             & mAP     & FPS        \\ \hline
				\textit{\textbf{One-stage}} & \multicolumn{17}{c}{}                                                                                                                                                                                                                                                                    \\ \hline
				PIoU~\cite{chen2020piou}                        & DLA-34    & 80.90          & 69.70          & 24.10          & 60.20          & 38.30          & 64.40          & 64.80          & \textbf{90.90} & 77.20          & 70.40          & 46.50          & 37.10          & 57.10          & 61.90          & 64.00          & 60.50       & -     \\
				RetinaNet-O*~\cite{lin2017focal}              & R-50-FPN  & 88.67          & 77.62          & 41.81          & 58.17          & 74.58          & 71.64          & 79.11          & 90.29          & 82.18          & 74.32          & 54.75          & 60.60          & 62.57          & 69.67          & 60.64          & 68.43       & -   \\
				DRN~\cite{pan2020dynamic}                         & H-104     & 88.91          & 80.22          & 43.52          & 63.35          & 73.48          & 70.69          & 84.94          & 90.14          & 83.85          & 84.11          & 50.12          & 58.41          & 67.62          & 68.60          & 52.50          & 70.70       & -   \\ 
				DAL~\cite{ming2020dynamic}                         & R-50-FPN  & 88.68          & 76.55          & 45.08          & 66.80          & 67.00          & 76.76          & 79.74          & 90.84          & 79.54          & 78.45          & 57.71          & 62.27          & 69.05          & 73.14          & 60.11          & 71.44       & -   \\ 
				RSDet~\cite{qian2019learning}                       & R-101-FPN & 89.80          & 82.90          & 48.60          & 65.20          & 69.50          & 70.10          & 70.20          & 90.50          & 85.60          & 83.40          & 62.50          & 63.90          & 65.60          & 67.20          & 68.00          & 72.20      & -    \\ 
				R3Det~\cite{yang2019r3det}                       & R-101-FPN & 88.76          & 83.09          & 50.91          & 67.27          & 76.23          & 80.39          & 86.72          & 90.78          & 84.68          & 83.24          & 61.98          & 61.35          & 66.91          & 70.63          & 53.94          & 73.79       & 12.1   \\ 
				S$^2$ANet~\cite{han2021}                      & R-50-FPN  & 89.11          & 82.84          & 48.37          & 71.11          & 78.11          & 78.39          & 87.25          & 90.83          & 84.90          & 85.64          & 60.36          & 62.60          & 65.26          & 69.13          & 57.94          & 74.12       & 15.3   \\ \hline
				\textit{\textbf{Two-stage}} & \multicolumn{17}{c}{}                                                                                                                                                                                                                                                                    \\ \hline
				RRPN~\cite{ma2018}                        & R-101     & 80.94          & 65.75          & 35.34          & 67.44          & 59.92          & 50.91          & 55.81          & 90.67          & 66.92          & 72.39          & 55.06          & 52.23          & 55.14          & 53.35          & 48.22          & 60.01      & -    \\ 
				R2CN~\cite{jiang2017r2cnn}                       & R-101     & 80.94          & 65.67          & 35.34          & 67.44          & 59.92          & 50.91          & 55.81          & 90.67          & 66.92          & 72.39          & 55.06          & 52.23          & 55.14          & 53.35          & 48.22          & 60.67      & -    \\ 
				RoI Transformer~\cite{ding2019}             & R-101-FPN & 88.64          & 78.52          & 43.44          & 75.92          & 68.81          & 73.68          & 83.59          & 90.74          & 77.27          & 81.46          & 58.39          & 53.54          & 62.83          & 58.93          & 47.67          & 69.56     & 11.3     \\ 
				SCRDet~\cite{yang2019}                      & R-101-FPN & \textbf{89.98} & 80.65          & 52.09          & 68.36          & 68.36          & 60.32          & 72.41          & 90.85          & \textbf{87.94} & 86.86          & 65.02          & 66.68          & 66.25          & 68.24          & 65.21          & 72.61      & -    \\ 
				RoI Transformer*~\cite{ding2019}            & R-50-FPN  & 88.65          & 82.60          & 52.53          & 70.87          & 77.93          & 76.67          & 86.87          & 90.71          & 83.83          & 82.51          & 53.95          & 67.61          & 74.67          & 68.75          & 61.03          & 74.61     & 11.3     \\ 
				Gliding Vertex~\cite{xu2020}             & R-101-FPN & 89.64          & 85.00          & 52.26          & 77.34          & 73.01          & 73.14          & 86.82          & 90.74          & 79.02          & 86.81          & 59.55          & 70.91          & 72.94          & 70.86          & 57.32          & 75.02     & -     \\ 
				Faster RCNN-O*~\cite{ren2016}           & R-50-FPN  & 88.44          & 73.06          & 44.86          & 59.09          & 73.25          & 71.49          & 77.11          & 90.84          & 78.94          & 83.90          & 48.59          & 62.95          & 62.18          & 64.91          & 56.18          & 69.05      & 14.9    \\  \hline
				\textit{\textbf{Ours}} & \multicolumn{17}{c}{}                                                                                                                                                                                                                                                                    \\ \hline
				AOPG                   & R-50-FPN  & 89.27          & 83.49          & 52.50          & 69.97          & 73.51          & 82.31          & 87.95          & 90.89          & 87.64          & 84.71          & 60.01          & 66.12          & 74.19          & 68.30          & 57.80          & 75.24      & 12.4    \\ 
				AOPG                    & R-101-FPN & 89.14          & 82.74          & 51.87          & 69.28          & 77.65          & 82.42          & 88.08          & 90.89          & 86.26          & 85.13          & 60.60          & 66.30          & 74.05          & 67.76          & 58.77          & 75.39     & 12.4     \\ 
				AOPG$^\ddagger$                   & R-50-FPN  & 89.88          & 85.57          & \textbf{60.90} & \textbf{81.51} & \textbf{78.70} & \textbf{85.29} & \textbf{88.85} & 90.89          & 87.60          & \textbf{87.65} & \textbf{71.66} & 68.69          & 82.31          & \textbf{77.32} & \textbf{73.10} & \textbf{80.66}   & 10.8  \\ 
				AOPG$^\ddagger$                  & R-101-FPN & \textbf{89.98} & \textbf{86.14} & 60.20          & 79.55          & 78.47          & 84.93          & 88.79          & 90.88          & 87.32          & 87.07          & 71.50          & \textbf{71.22} & \textbf{83.57} & 72.47          & 70.77          & 80.19       & 10.8   \\ \hline
		\end{tabular}}
		\vspace{-0.3cm}
	\end{center}
\end{table*}

\begin{table*}\small
	\setlength{\abovecaptionskip}{-0.2cm}
	\setlength{\belowcaptionskip}{-0.2cm}
	\caption{Comparison with state-of-the-art methods on the HRSC2016 dataset.}
	\vspace{-0.2cm}
	\label{table:hrsc_comparison}
	\begin{center}
		\renewcommand\arraystretch{1.5}
		\resizebox{1.0\textwidth}{!}{
			\begin{tabular}{p{2.2cm}<{\centering}|p{1.7cm}<{\centering}|p{1.7cm}<{\centering}|p{1.7cm}<{\centering}|p{1.7cm}<{\centering}|p{1.7cm}<{\centering}|p{1.7cm}<{\centering}|p{1.7cm}<{\centering}|p{1.7cm}<{\centering}|p{1.7cm}<{\centering}}
				\hline
				\multicolumn{1}{c|}{ } & RRPN~\cite{ma2018}  & R2CNN~\cite{jiang2017r2cnn} & RT~\cite{ding2019} & GV~\cite{xu2020}    & DRN~\cite{pan2020dynamic}   & PIoU~\cite{chen2020piou}  & DAL~\cite{ming2020dynamic}   & S$^2$ANet~\cite{han2021} & \textbf{AOPG}  \\ 
				\hline
				mAP (VOC 07)                           & 79.08 & 73.07 & 86.20     & 88.20 & -     & 89.20 & 89.77 & 90.17  & \textbf{90.34} \\ 
				mAP (VOC 12)                           & 85.64 & 79.73 & -         & -     & 92.70 & -     & -     & 95.01  & \textbf{96.22} \\ \hline
		\end{tabular}}
	\end{center}
	\vspace{-0.2cm}
\end{table*}

\subsection{Comparison with State-of-the-Art Methods}

\textbf{Results on the DIOR-R dataset.} We test several advanced oriented detectors on the DIOR-R dataset and list their results in Table~\ref{table:dior_comparison}. All experiments use the trainval set in training and evaluate on the test set. As the experiments show, the Faster R-CNN OBB and RetineNet OBB can barely reach 59.54$\%$ and 57.55$\%$ mAP. The improvement methods, such as Gliding Vertex and RoI Transformer, only achieve 60.06$\%$ mAP and 63.87$\%$ mAP on the DIOR-R dataset. This demonstrate our DIOR-R is challenging for recent oriented detectors and has a huge room for growth. Compared with other methods, AOPG achieves dominant accuracy of 64.41$\%$ mAP, which verifies the robustness and generalization of our methods. We visualize the detection results in Fig.~\ref{img:dior_vis}.

\textbf{Results on the DOTA dataset.} We report the results of 14 oriented detectors in Table~\ref{table:dota_comparison}. Without any bells and whistles, our AOPG achieves 75.24$\%$ mAP based on ResNet50-FPN and 75.39$\%$ mAP based on ResNet101-FPN, which has surpassed other advanced oriented detection methods. With the multi-scale training and testing and rotated augmentation, we reach 80.66$\%$ mAP with ResNet50-FPN as backbone and 80.19$\%$ mAP with ResNet101-FPN as backbone, which are the highest results on the DOTA dataset. Our models also achieve the best results in some very challenging categories, such as bridge, soccer-ball field and large-vehicle. We visualize some detection results and show them in Fig.~\ref{img:dota_vis}.

In the aspect of speed, AOPG can achieve 12.4 FPS on the DOTA dataset with the ResNet50 backbone, which is only 2.5 FPS slower than the original Faster R-CNN OBB. However, AOPG can boost the mAP by a big margin of 6.19\%. This shows that our AOPG can largely improve the oriented object detection accuracy with acceptable time consumption.

\begin{figure*}
	\begin{center}
		\vspace{-0.1cm}
		\includegraphics[width=0.95\linewidth]{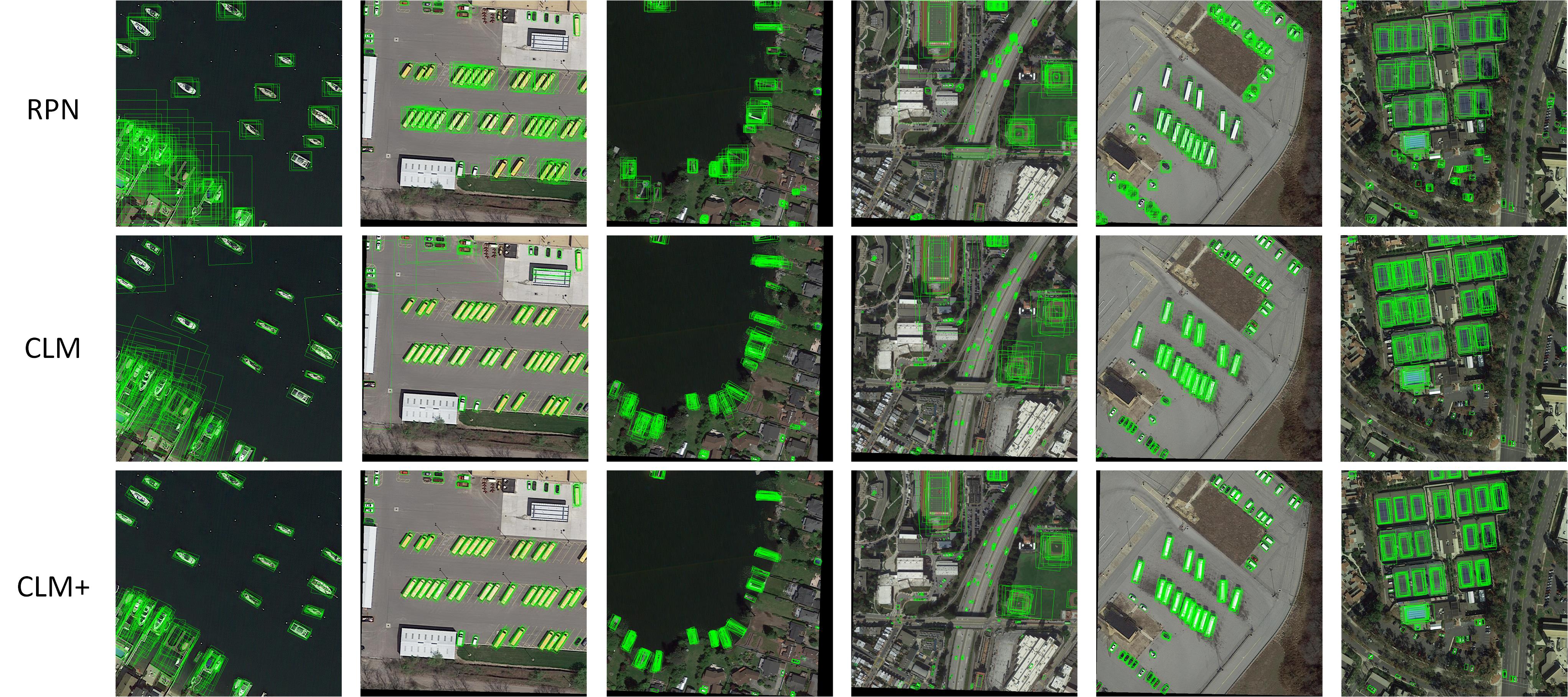}\\
	\end{center}
	\vspace*{-0.4cm}
	\caption{Proposals generated with different methods including RPN, CLM and CLM+.}\label{img:clm_comparison}
\end{figure*}

\textbf{Results on the HRSC2016 dataset.} The results of the HRSC2016 dataset are shown in Table~\ref{table:hrsc_comparison}. We use both PASCAL VOC07 and VOC12~\cite{everingham2010pascal} metrics to evaluate our model. As shown, our AOPG achieves 90.34$\%$ and 96.22$\%$ mAP, when using the VOC07 metric and the VOC12 metric, respectively, both of which are very competitive compared with current state-of-the-art methods. We visualize the detection results in Fig.~\ref{img:hrsc_vis}.

\begin{table}[]
\caption{Comparison of different oriented proposal generators (OPGs) on the DOTA validation set.}
\label{table:clm_comparison}
\renewcommand{\arraystretch}{1.2}
\resizebox{0.49\textwidth}{!}{
\begin{tabular}{m{0.7cm}<{\centering}m{1.5cm}<{\centering}m{1.3cm}<{\centering}m{0.7cm}<{\centering}|m{1.2cm}<{\centering}|m{1cm}<{\centering}m{1cm}<{\centering}}
\hline
\multicolumn{4}{c|}{Oriented Proposal Generators (OPGs)}                           & \multirow{2}{*}[-5pt]{AlignConv}  & \multicolumn{1}{c}{\multirow{2}{*}[-5pt]{Recall (\%)}} & \multirow{2}{*}[-5pt]{mAP (\%)} \\ \cline{1-4}
RoI-Trans     & Anchor-based OPG & FCOS-based OPG & CLM        &                            & \multicolumn{1}{c}{}                             &                           \\ \hline
\checkmark      &                        &                      &            &                            & 87.86                                            & 73.12                     \\
                & \checkmark             &                      &            &                            & 86.14                                            & 72.53                     \\
                & \checkmark             &                      &            & \checkmark                 & 86.96                                            & 73.09                     \\
                &                        & \checkmark           &            &                            & 85.44                                            & 72.17                     \\
                &                        & \checkmark           &            & \checkmark                 & 86.32                                            & 72.69                     \\
                &                        &                      & \checkmark &                            & 89.82                                            & 73.69                     \\
                &                        &                      & \checkmark & \checkmark                 & \textbf{90.68}                                   & \textbf{74.16}            \\ \hline
\end{tabular}}
\vspace*{-0.4cm}
\end{table}

\subsection{Comparisons of Different Oriented Proposal Generators}

We conduct experiments to compare our AOPG with other oriented proposal generators (OPGs). The experiments are conducted on the DOTA validation set and the results are reported in Table~\ref{table:clm_comparison}. Here, RoI-Trans indicates applying RoI Learner in RoI Transformer~\cite{ding2019} as the proposal generator. Anchor-based OPG means using horizontal anchors and directly regressing the angles in the proposal generation phase. FCOS-based OPG means applying the FCOS regression method when generating oriented proposals.

As we can see in Table~\ref{table:clm_comparison}, RoI-Trans, anchor-based OPG, and FCOS-based OPG can only reach 73.12\%, 72.53\%, and 72.17\% mAP respectively. After applying AlignConv, their accuracies are boosted to 73.09\% and 72.69\%. On the contrast, our CLM can achieve 73.69\% mAP on its own and 74.16\% with AlignConv, which surpasses other OPG methods. We also compare the Recall of different proposal generation methods on the DOTA validation set. Our CLM achieves 89.82\% on its own and up to 90.68\% with AlignConv. Those results confirm the motivation of AOPG that relying on horizontal boxes will harm the oriented object prediction. Our AOPG is in anchor-free manner and gets rid of horizontal boxes so that it outperforms other methods. What is more, the refinement module (AlignConv) boosts the Recall gains of anchor-based OPG, FCOS-based OPG, and CLM with the margins of 0.82, 0.88, and 0.86 points, respectively, and also respectively improves the mAP gains with the margins of 0.56, 0.52, and 0.47 points. The results demonstrate that the refinement module can indeed further improve the quality of the proposals and benefit the oriented detection.

To further illustrate the effectiveness, we visualize the top 300 proposals generated by traditional RPN and the different parts of AOPG. The comparison results are given in Fig.~\ref{img:clm_comparison}. Here, CLM+ indicates CLM with AlignConv. We can see that the horizontal proposals generated by RPN are confused at the places where the objects are densely packed. For instance, the horizontal proposals cannot separate harbors and large-vehicles in the images. In contrast, the oriented proposals generated by the CLM can well localize them and also better fit the ground truths than horizontal proposals. Furthermore, the CLM+ produces more accurate oriented proposals and eliminates some false positive samples generated by CLM.

\begin{table}[]
\caption{valuations of AOPG with different frameworks.}
\label{table:framework}
\renewcommand{\arraystretch}{1.2}
\begin{tabular}{m{0.7cm}<{\centering}|m{2.1cm}<{\centering}|m{2.1cm}<{\centering}|m{2.1cm}<{\centering}}
\hline
     & Faster R-CNN                     & Cascade R-CNN                     & Double Head                     \\ \hline
RPN  & 69.05                            & 73.45                             & 72.03                           \\
AOPG & 75.24                            & 76.31                             & 75.56                           \\ \hline
\end{tabular}
\vspace*{-0.2cm}
\end{table}

\subsection{Evaluations of AOPG with Different Frameworks}

To further boost the performance and evaluate the consistency of the effectiveness of AOPG, we supplement the experiments based on more advanced frameworks like Cascade R-CNN~\cite{cai2018cascade} and Double Head~\cite{wu2019rethinking}. The results are reported in Table~\ref{table:framework}. As we can see, the Cascade R-CNN and Double Head can only achieve 73.45\% and 72.03\% mAP. However, after replacing the original RPN with AOPG, their accuracies measured in terms of mAP increase to 76.31\% (2.86$\uparrow$) and 75.56\% (3.53$\uparrow$), which displays the consistent effectiveness of our work.

\section{Conclusion}
In this paper, we analyzed the drawbacks of using horizontal boxes for oriented object detection and concluded that involving horizontal boxes into object detectors will hurt the final results. Taking this problem as a starting point, we proposed a novel Anchor-free Oriented Proposal Generator (AOPG) that totally removes all horizontal boxes in the network. Moreover, we released a new large-scale oriented object detection dataset, named DIOR-R, to alleviate the shortage of data. Comprehensive experiments show that our AOPG contributes to the increases of detection accuracy. Specifically, our AOPG achieves the highest accuracy without any bells and whistles and obtains large gains over the baseline on all DIOR-R, DOTA and HRSC2016 datasets.

\ifCLASSOPTIONcaptionsoff
  \newpage
\fi

\bibliographystyle{IEEEtran}
\bibliography{reference}

\end{document}